\documentclass[11pt]{article}
\usepackage[margin=1in]{geometry}
\usepackage{amsmath,amssymb,booktabs,graphicx,natbib,hyperref,xcolor}
\graphicspath{{figures/}}
\hypersetup{colorlinks=true, linkcolor=black, citecolor=blue, urlcolor=blue, pdftitle={Is the Platform Part of the Measurement? A Protocol and Simulation Study for Cross-Platform Equivalence of Machine-Labelled Policy Sentiment}, pdfauthor={Asad Ikram and Ella Haig}}

\title{Is the Platform Part of the Measurement?\\ \Large A Protocol and Simulation Study for Cross-Platform Equivalence of Machine-Labelled Policy Sentiment}
\author{Asad Ikram\thanks{Independent researcher. This work began as the first author's MSc dissertation at the University of Portsmouth (2025). Correspondence: asad.ikram53@gmail.com} \and Ella Haig\thanks{School of Computing, Mathematics \& Physics, University of Portsmouth, UK}}
\date{October 2026}

\begin{document}
\maketitle
\begin{abstract}
Comparing machine-labelled sentiment across social media platforms is
now routine: differences in classifier scores between platforms are
read as differences in public opinion. These comparisons rest on an
assumption that the classifier measures the same construct, on the
same scale, on every platform. Some studies check a classifier's validity platform by platform, but to our knowledge none has modelled the platform as a facet of the classifier's measurement. We adapt the psychometric
equivalence toolkit (generalizability theory, measurement invariance
and differential item functioning, and score linking) to
machine-labelled policy sentiment, treating the platform as a survey
mode, and specify a six-step protocol with anchor events fixed in advance, an anchor-density audit, and a two-classifier convergence
control. A simulation with known ground truth illustrates what the assumption can cost in one generating configuration fixed in advance: across 500 replications, because the platforms measure through different functions, a pooled index that takes platform scores at face value reports the wrong sign of aggregate sentiment on nearly a quarter of days, whereas an index adjusted with each platform's known measurement function does so on under one day in sixty on average. A case study of the instrument from the first author's MSc dissertation, a fine-tuned BERT classifier over 255,446 UK
economic-policy comments from Twitter/X, Reddit, YouTube, and Quora,
motivates the design and illustrates the instrument-dependence the protocol addresses. The full four-platform application, under fresh collection with the pre-specified design, is the designated follow-up study. The protocol lets a researcher test whether pooling
platforms is defensible in a given corpus, or whether platform scores
must be linked before any average is taken.
\end{abstract}

\section{Introduction}
\label{sec:intro}

Computational social science now compares machine-labelled text across platforms as a matter of course. \citet{overgaard2024} apply a BERT
classifier to Facebook and Twitter posts and compare the resulting
scores directly. \citet{wei2026} run the same
design across Douyin and TikTok. Studies in this mould publish a
cross-platform comparison, or a pooled multi-platform index, on the
assumption that the classifier behaves identically wherever it is
pointed. The assumption is rarely stated, still more rarely examined, and it is not an obviously safe one. \citet{nogara2025} report that Perspective API scores German comments as more toxic than equivalent English ones: the instrument behaves differently in different languages. In \citet{atreja2025}, LLM annotations shift under changes of prompt wording alone. And \citet{greene2025} find that the same members of Congress present as more ideologically extreme on Facebook than on Twitter, judged by the news domains they share: the same respondents, measured in two modes, give different readings. Classifiers, like other text-derived measures, are instruments, and their behaviour depends on the conditions in which they are used.

\begin{figure}[t]
\centering
\includegraphics[width=0.8\linewidth]{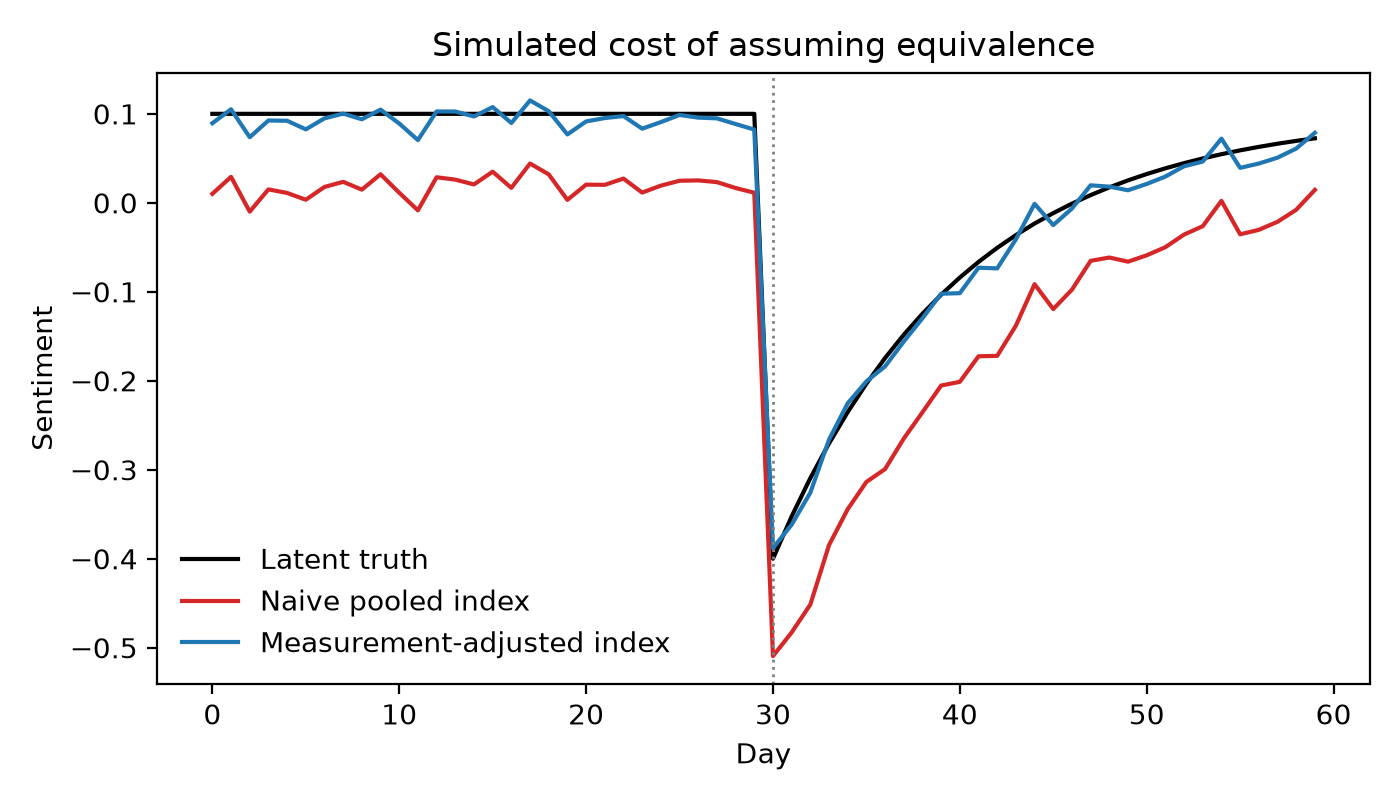}
\caption{What assuming equivalence costs, in one simulation run with known ground truth (seed 20261201; parameters fixed ex ante and released with the code; Appendix~\ref{app:availability}). The naive pooled index reports the wrong sign of aggregate sentiment on 13 of 60 days in this run; the measurement-adjusted index tracks the latent truth with no wrong sign day. Here the true event shift is $-0.332$; the naive index estimates $-0.369$ and the adjusted index $-0.326$; the naive shift error mainly reflects the coarse three point scale rather than non-equivalence (Section~\ref{sec:cost}). Section~\ref{sec:cost} reports 500 replications.}
\label{fig:simdemo}
\end{figure}

When the labels are machine-produced, whether the platform is part of the measurement is rarely tested formally. When a
classifier scores Reddit discussion of a budget more negative than
tweets about the same budget, the difference is read as a fact about opinion, when it could equally be a property of the instrument. Character
limits, anonymity, threading, moderation, and audience shape how the
same attitude is expressed, and therefore how it is scored.
\citet{sen2021} give this hazard a name, platform-affordance error,
in their total error framework for digital traces, but they provide no
formal test for it. Recent correction methods repair label error
against a human gold standard \citep{egami2023} and probe the
fragility of LLM annotation pipelines
\citep{baumann2025,halterman2025}, yet none of them models the
platform as a source of measurement error.

Survey methodology met a structurally identical problem decades ago
and built machinery for it. The same respondent answers differently by
telephone, on paper, and face to face; the field calls this a mode
effect and does not treat cross-mode differences as substantive until
measurement equivalence has been established
\citep{tourangeau2000,klausch2013,hox2015}, within a total-error
accounting of where distortions enter \citep{groves2010}. Our claim is that a platform is a mode. The analogy has limits: absent random assignment of users to platforms or within-user cross-posting, the design identifies platform-level non-equivalence, a bundle of selection and measurement that Section~\ref{sec:limitations} unpacks, with mode as the leading interpretation rather than an established fact. Affordances stand to posts as interview
settings stand to answers, and the psychometric toolkit built for modes and groups (generalizability theory \citep{cronbach1972,brennan2001}, measurement invariance testing \citep{meredith1993,putnick2016}, and score linking \citep{kolen2014}) transfers to machine-labelled constructs with the platforms as the groups. The toolkit has reached survey self-reports \citep{boyd2024},
annotator facets \citep{kennedy2020,sachdeva2022}, and single-platform
text scales \citep{pokropek2026}, but the platform itself has stayed
an unmodelled stratum.

The question is also practical. Since the closure of the major
research APIs, few teams can study one platform at depth; instead they
assemble patchworks of whatever platforms remain accessible, with
heterogeneous coverage and access terms
\citep{davidson2023,goanta2026}. Every such patchwork implicitly pools
or compares scores across platforms. If a unit of classifier score
does not mean the same thing on Reddit as on X, the pooled trend lines
and cross-platform contrasts now filling the literature carry an error term of unknown size.

This paper makes four contributions.
\begin{enumerate}
  \item We formalise the platform-as-mode question for machine-labelled
    constructs and adapt the
    psychometric equivalence toolkit to machine-classified policy
    sentiment with platform as the grouping facet.
  \item We specify a six-step protocol (Sections~\ref{sec:divergence} to~\ref{sec:cost}): same-event divergence testing, a fixed-facet generalizability study against a human-validated criterion, categorical invariance and DIF under effect-size criteria, multilevel adjustment for composition and timing, anchor-based linking centred on the September 2022 mini-budget, and a simulation of the cost of skipping the first five steps. Every sentiment-indicator test is replicated with a non-lexicon transformer, so only convergent non-invariance is read as platform-level non-equivalence rather than a classifier artefact \citep{czarnek2022}, and invariance results are treated as diagnostic evidence rather than grounds to pool \citep{robitzsch2023}.
  \item We quantify the cost of assuming equivalence in a simulation with known ground truth (Section~\ref{sec:cost}, Figure~\ref{fig:simdemo}): across 500 replications of one generating configuration fixed in advance, a naive pooled index, because the platforms measure through different functions, reports the wrong sign of aggregate sentiment on nearly a quarter of days, an error that an index adjusted with the known measurement functions all but avoids.
  \item We ground the protocol in the instrument from the first author's MSc dissertation (Section~\ref{sec:data}), a fine-tuned BERT classifier over 255,446 UK economic-policy comments from Twitter/X, Reddit, YouTube, and Quora, whose execution records supply the motivating evidence; the full four-platform application, under fresh collection with the pre-specified design, is the designated follow-up study.
\end{enumerate}

\section{Related Work}
\label{sec:related}

\paragraph{Mode effects in survey measurement.}
Survey methodology settled long ago that the instrument is part of the measurement. Answers to the same question differ by mode of administration because each mode changes the cognitive and social conditions of answering \citep{tourangeau2000}. \citet{klausch2013} showed that interviewer-administered (face-to-face, telephone) and self-administered (paper, web) modes measure identical attitude scales non-equivalently, although modes within each pair were equivalent, and mixed-mode designs now model mode explicitly instead of ignoring it as a nuisance \citep{hox2015}. The Total Survey Error framework gives these effects a named place in the error budget \citep{groves2010}. \citet{sen2021} transported this thinking to digital traces and explicitly identified platform-affordance error as the trace analogue of mode error, but the framework names the error without supplying a test for it. Our premise is simply that platform should receive the same formal treatment for machine-labelled trace data that mode receives for surveys.

\paragraph{Measurement for machine-labelled text.}
A second literature treats classifiers as measurement instruments. \citet{jacobs2021} argue that computational pipelines operationalise unobservable constructs through measurement models whose validity must be tested, and \citet{halterman2025} extend this framing to large language models. Empirical work in this vein audits the instrument itself. \citet{pellert2022} validate sentiment macroscopes against self-report. \citet{atreja2025} show that model-generated labels shift with prompt design in unpredictable ways, and \citet{baumann2025} show that defensible configuration choices alone can flip downstream conclusions. \citet{egami2023} correct downstream estimates for classifier error with design-based supervised learning, but the correction targets the gold-standard label, so it can absorb classifier error that differs by platform yet leaves in place a platform effect on how sentiment is expressed, which human coders read too. The quantification literature corrects prevalence estimates for misclassification, either by adjusting classify-and-count with estimated error rates \citep{forman2008} or by estimating category proportions directly without classifying individual documents \citep{hopkins2010}, and allows those proportions to differ between labelled and target data; our adjusted index is close in spirit to adjusted classify-and-count applied platform by platform, but both approaches assume that the classifier's error rates, or the distribution of text within each category, are the same in the labelled and target data, which is the assumption a platform-level test examines. In all of this work the classifier is scrutinised; the platform it reads from is not.

\paragraph{Cross-platform studies and assumed equivalence.}
Multi-platform designs have become standard, and the post-API turn has made heterogeneous data regimes the norm \citep{davidson2023, goanta2026}. \citet{overgaard2024} apply a single BERT classifier across Facebook and Twitter and compare the resulting scores directly; \citet{wei2026} compare Douyin and TikTok in the same way, and \citet{singh2024} compare toxicity scores between Reddit and Discord communities. There is growing evidence that this is risky. \citet{greene2025} find that the same members of Congress present as more ideologically extreme on Facebook than on Twitter, and \citet{nogara2025} show that Perspective API scores German-language text as more toxic, a measurement artefact that masquerades as a substantive difference. Equivalence can also hold: \citet{liu2023} applied pre-trained Facebook-based language models to the Facebook posts and SMS messages of 120 users who shared both, and for six of eight models found no significant difference between the platforms in how well the estimates correlated with self-reports, so equivalence is an empirical question to be tested rather than assumed in either direction. The field's classic cautions concern populations and samples: who is on the platform \citep{tufekci2014}, how the stream is sampled \citep{morstatter2013}, whether models generalise beyond their training population \citep{cohen2013}, and whether text sentiment tracks opinion at all \citep{oconnor2010}. The complementary question, whether the same score means the same thing on Reddit as on YouTube, is routinely assumed and almost never tested.

\paragraph{The psychometric toolkit and nearest misses.}
The tools for this question already exist in psychometrics. Generalizability theory partitions score variance among measurement facets \citep{cronbach1972, brennan2001}; measurement invariance testing asks whether indicators relate to a construct identically across groups \citep{meredith1993, putnick2016}; linking places scores from different instruments on a common scale \citep{kolen2014}; and invariance results are read as diagnostic evidence rather than permission to compare \citep{robitzsch2023}. Several recent studies come close to combining this toolkit with platform data without quite doing so. \citet{boyd2024} do test cross-platform invariance, though of survey self-reports rather than machine labels. Faceted Rasch measurement has reached annotation through \citet{kennedy2020} and \citet{sachdeva2022}, with annotators as the modelled facet. In \citet{pellert2022} validation is by correlation alone, with no invariance structure. The multitrait-multimethod models of \citet{cernat2025} compare survey with digital-trace measures rather than platform with platform. \citet{han2025} document platform-dependent selection and measurement bias, but in online review ratings rather than machine-labelled text. \citet{pokropek2026} fit factor models to text indicators while grouping by time within a single platform. And generalizability designs have reached LLM annotation, with prompt and model choice as variance facets, without yet reaching platform \citep{camuffo2026}. Rasch and error-decomposition frameworks have likewise reached LLM raters and evaluation pipelines \citep{sachdeva2026,messing2026}, with the rater or the pipeline configuration as the modelled facet.

To our knowledge, no study has applied the psychometric equivalence toolkit of generalizability, invariance, and linking to machine-classified policy sentiment with platform as the grouping facet, and this paper specifies that test.

\section{The Motivating Case and the Pre-Specified Application}
\label{sec:data}

This section describes the corpus and instrument from the first author's MSc dissertation \citep{ikram2025} that motivate the paper, and the validation design the pre-specified application will execute. The account rests on the dissertation's execution records (its code, pipeline logs, and fully executed notebook); Section~\ref{sec:limitations} documents where the dissertation text is internally inconsistent or is not reproduced by those records. Dissertation-stage figures are reproduced in their original form throughout, as evidence of the instrument as it stood; Appendix~\ref{app:diagnostics} states the figure standard the application will apply to its fresh counterparts.

\paragraph{Corpus.}
The dissertation-stage corpus comprised 255,446 public comments attached to 6,650 parent posts, about UK economic policy (per the pipeline records; the dissertation text gives approximately 279,000, a figure the surviving records do not reproduce). It was collected from Twitter/X (licensed third-party archive, January 2018 to July 2025; dissertation p.~21), Reddit (official developer API, no stated date restriction; p.~22), YouTube (Data API v3, top-level comments on major UK news and government channels; pp.~22-23), and Quora (the same licensed analytics partner as Twitter; p.~23). The precise collection window cannot be reconstructed: the dissertation scopes the study to 2010 to mid-2025 (p.~7), gives January 2018 to July 2025 for Twitter only (p.~21), and states no restriction for Reddit (p.~22); the application fixes its window ex ante instead.

\begin{figure}[t]
\centering
\includegraphics[width=\linewidth]{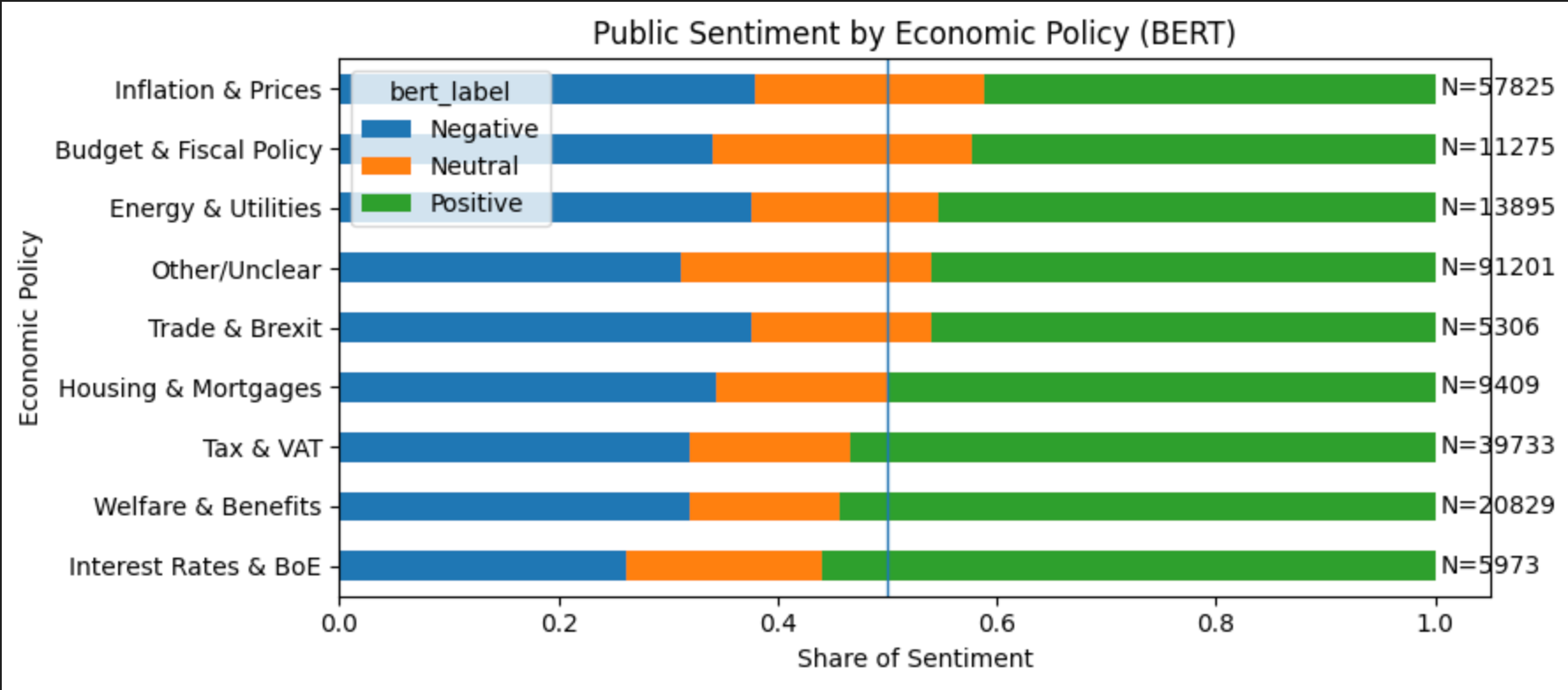}
\caption{Sentiment shares by policy category under the fine-tuned BERT classifier, as they stood at dissertation stage (Figure~7.12, p.~41). The nine keyword-assigned categories carry per-category counts that sum to 255,446, the total the surviving pipeline records establish, and the Other/Unclear category absorbs roughly a third of classified posts.}
\label{fig:policy}
\end{figure}

Collection was organised by an auditable keyword scheme with six seed-rule topics: taxation and fiscal policy; inflation, cost of living and prices; welfare and benefits; energy and utilities; housing and mortgages; and the labour market (dissertation p.~18; the scheme, including its disambiguation priority and negation window, is transcribed in Appendix~\ref{app:keywords}). Keyword tagging then assigned classified posts to nine analysis categories, of which an Other/Unclear category absorbed roughly a third (dissertation pp.~41-42). Per-category counts are recorded in the executed figure (the N labels of Figure~\ref{fig:policy}). The unit of analysis is the comment, with parent posts available as a clustering level. The execution records do not separate the corpus by platform: the dissertation's per-platform figures (approximately 120k Twitter, 130k Reddit, 120k YouTube, 25k Quora; pp.~21-24) sum to roughly 395k, which the records do not allow us to reconcile with either the 279k figure or the 255,446 pipeline total. The application reports its composition from collection logs by construction.

Machine labels came from a fine-tuned BERT classifier \citep{devlin2019} (macro-F1 0.878 against corpus labels seeded by VADER, a rule-based sentiment lexicon \citep{hutto2014}, not against the human criterion, computed over the full corpus, of which 90 per cent was training data; dissertation Table~7.1, p.~35), trained on VADER-seeded labels; Table~\ref{tab:models} reproduces the dissertation-stage comparison and Figure~\ref{fig:policy} the per-category sentiment shares it produced. We disclose this circularity and address it with the human criterion below and with the artefact control in Section~\ref{sec:framework}. The dissertation-stage shares themselves fail face validity: every policy category is plurality positive for UK economic discourse spanning a cost-of-living crisis, which is substantively implausible and is best read as a candidate artefact of the lexicon-seeded labels. The expanded human criterion measures this directly; if it confirms a positive bias, that bias enters the generalizability and invariance analyses as a measured quantity rather than an assumption. The application holds its corpus fixed across every corpus-based step, so every comparison uses the same comments.

\begin{table}[t]
\centering
\caption{Dissertation-stage classifier comparison (dissertation Table~7.1, p.~35). The BERT and BiLSTM rows match the appendix confusion matrices exactly and therefore measure agreement with the VADER-seeded labels over the full classified corpus; the reference standard for the VADER and logistic regression rows is not recorded, and their accuracies are not whole-number proportions of 200, so they cannot be accuracies over the full 200-post human set. The rows therefore cannot be assumed to share a reference standard, and no row measures fine-tuned performance against the human criterion; the dissertation's own notebook output computes the BERT figures against the corpus labels (p.~31). Values are shown as printed in the dissertation. The application will retrain BERT and the BiLSTM by the procedure the executed code records on its own corpus and score every model against both reference standards, with binomial intervals on the human-criterion figures.}
\label{tab:models}
\begin{tabular}{lcccc}
\toprule
Model & Accuracy & Macro precision & Macro recall & Macro F1 \\
\midrule
BERT (fine-tuned) & 88.06\% & 0.8815 & 0.8744 & 0.8778 \\
BiLSTM (GloVe) & 81.51\% & 0.8141 & 0.8099 & 0.8111 \\
VADER baseline & 65.20\% & 0.66 & 0.65 & 0.65 \\
Logistic regression (BoW) & 71.40\% & 0.72 & 0.71 & 0.71 \\
\bottomrule
\end{tabular}
\end{table}

\paragraph{Anchor events.}
Cross-platform comparison requires posts about the same event at the same time. We fixed fifteen UK fiscal policy events (Table~\ref{tab:anchors}) on 21 August 2026, before any window count was computed, with the September 2022 mini-budget as the centrepiece, and defined two windows per event: an acute 48-hour window and an extended two-week window. The application's event-assignment rule is fixed ex ante: the keyword scheme of Appendix~\ref{app:keywords} applied to the comment text and parent title, with the platform's comment timestamp as the time field, and relevance validated on a manual subsample within the expanded annotation. We expect the acute windows to be sparse on Quora, whose question-and-answer format does not lend itself to event-locked measurement; the audit is designed to identify such platforms so they can be given a reduced role or excluded, and when a platform is excluded we report a gap in the data and claim nothing about its equivalence. An anchor-density audit precedes all analysis (Section~\ref{sec:framework}); platforms that fail the minimum post-count threshold for an event-window are excluded from that comparison, and the analysis is restricted to the platforms with adequate density if necessary. The dissertation reports lower data density in early 2022 (p.~25), so density is checked per event-window rather than assumed.

\begin{table}[t]
\centering
\caption{Anchor events fixed on 21 August 2026, before any window count was computed. Per-platform counts in the 48-hour and two-week windows are produced by the application's density audit on its corpus; the inclusion thresholds, fixed at the same time, are 200 comments per platform-event cell in the two-week window and 50 in the 48-hour window. The audit re-verifies each date against official records before counting.}
\label{tab:anchors}
\begin{tabular}{ll}
\toprule
Event & Date \\
\midrule
Budget 2020 (Covid response) & 11 Mar 2020 \\
Furlough scheme announced & 20 Mar 2020 \\
Budget 2021 & 3 Mar 2021 \\
Autumn Budget 2021 & 27 Oct 2021 \\
Energy bills package and Bank rate rise & 3 Feb 2022 \\
Cost-of-living package and windfall tax & 26 May 2022 \\
Energy Price Guarantee & 8 Sep 2022 \\
Mini-budget (centrepiece anchor) & 23 Sep 2022 \\
Statement reversing the mini-budget & 17 Oct 2022 \\
Autumn Statement 2022 & 17 Nov 2022 \\
Spring Budget 2023 & 15 Mar 2023 \\
Autumn Statement 2023 & 22 Nov 2023 \\
Spring Budget 2024 & 6 Mar 2024 \\
Autumn Budget 2024 & 30 Oct 2024 \\
Spring Statement 2025 & 26 Mar 2025 \\
\bottomrule
\end{tabular}
\end{table}

\paragraph{Gold-standard sample.}
The dissertation validated machine labels on 200 posts, balanced across the four platforms, all labelled by hand by the first author, who reviewed each post's VADER-assigned label (dissertation p.~25); the dissertation's Cohen's $\kappa \approx 0.78$ (p.~28) measures agreement between those hand labels and VADER, not between two annotators. Because that sample is small and was not drawn by probability sampling, the pre-specified application replaces it with a new sample of 1,000-1,500 comments drawn from its own corpus by stratified probability sampling, with strata defined by platform, anchor event, and predicted class, and inclusion probabilities retained for design-weighted estimates. A second annotator codes an overlap subset, targeted at a quarter to a third of the sample; disagreements are adjudicated and inter-annotator reliability is reported from the pre-adjudication ratings. This sample is the criterion for the generalizability study and for validating the machine labels; the invariance models themselves are fitted on the machine-labelled corpus within anchor windows, with the criterion validating their inputs rather than forming their estimation sample. Because the expanded criterion is labelled blind to the machine labels, it breaks the circularity of the VADER-seeded training labels, which the dissertation-stage set, labelled by reviewing VADER output, could not. Human validation remains essential for machine annotation \citep{pangakis2025, atreja2025}.

\paragraph{Ethics.}
The underlying study completed the University of Portsmouth online ethics review (School of Computing; Certificate of Ethics Review TETHIC-2025-111094, dissertation p.~60) and was classified as low risk (p.~15), with legitimate interests as the lawful basis under UK GDPR (p.~14). We analyse public posts only and apply data minimisation. We report aggregates, paraphrase any quoted material, make no attempt to identify users, and, in the application, share dehydrated identifiers rather than full text where platform terms permit. The dissertation-stage corpus is not redistributed (Appendix~\ref{app:availability}); user identifiers were hashed (p.~25).

\section{Step 1: Establishing Divergence}
\label{sec:divergence}

Before asking whether the platform shapes the measurement, we establish that there is something to explain. For each anchor event and window, we compute the distribution of machine-assigned sentiment labels per platform and compare platforms pairwise. Because the comparison is confined to a shared event window, it constrains topic and time far more tightly than whole-corpus comparison, though different sub-events may still dominate on different platforms within a window; Step 4 models the timing component of that composition, and the remainder stays within the population difference discussed in Section~\ref{sec:limitations}. We summarise differences in class proportions with Cram\'er's $V$ and cluster-bootstrap confidence intervals, resampling threads and, where hashed identifiers permit, users; chi-square statistics are reported, but because comments are clustered within threads and at this sample size significance alone is uninformative, interpretation rests on the effect sizes and their clustered intervals. Divergence at this step is only a precondition. It is consistent both with genuine differences in opinion and with platform-level non-equivalence in measurement; Steps 2 to 4 test for the second, which, as Section~\ref{sec:limitations} explains, still bundles population composition with measurement mode. Prior work stops at this step or assumes it away \citep{pellert2022, overgaard2024}.

This step executes in the pre-specified application, on its own corpus; the simulation of Section~\ref{sec:cost} demonstrates, with known ground truth, what is at stake when the equivalence question is not asked at all.

\section{Steps 2 to 4: Reliability, Invariance, Adjustment}
\label{sec:framework}

We treat the classifier as a measurement instrument and the platform as a candidate component of the measurement procedure \citep{jacobs2021, halterman2025}. The design proceeds in six steps: establish divergence (Section~\ref{sec:divergence}); estimate reliability per platform with a generalizability study; test measurement invariance and differential item functioning with platform as the grouping facet; adjust for composition and timing in a multilevel model; link scales where the evidence permits (Section~\ref{sec:linking}); and quantify the cost of skipping the first five steps (Section~\ref{sec:cost}).

Two design commitments apply to every step that follows. The anchor-density audit comes first: every event-window-platform cell must meet a minimum post count before it enters any model, and if cells fail, the study narrows to the dense subset of platforms rather than papering over sparsity \citep{morstatter2013}. And platform is a fixed facet throughout: our claims concern these four platforms, not a universe of platforms from which they were sampled.

\subsection{Step 2: Fixed-facet generalizability study}
The criterion is the gold-standard sample of Section~\ref{sec:data}. For each platform $m$ we estimate a generalizability study with posts nested in events and coders as a random facet, analysed on pre-adjudication ratings; the design is unbalanced by construction, since the second coder rates a subset, and components are estimated by REML with interval estimates rather than point comparisons. The classifier enters the same decomposition as an additional fixed rater, and machine-human disagreement is estimated as a separate per-platform component alongside the coder decomposition:
\begin{equation}
X^{(m)}_{p(e)c} = \mu^{(m)} + \nu^{(m)}_{e} + \nu^{(m)}_{p(e)} + \nu^{(m)}_{c} + \nu^{(m)}_{ec} + \nu^{(m)}_{p(e)c,\varepsilon},
\label{eq:gstudy}
\end{equation}
with variance components for events, posts within events, coders, their interactions, and residual estimated separately at each fixed level $m$ \citep{cronbach1972, brennan2001}. Comparing absolute error components across platforms shows whether the measurement procedure is equally dependable on every platform, something a single reliability coefficient cannot show. Dependability coefficients are population-bound, so cross-platform coefficient comparisons are reported as descriptive only. Generalizability designs have recently modelled prompts and models as facets of automated annotation \citep{camuffo2026}, and annotator-facet studies exist \citep{kennedy2020, sachdeva2022}, but none of them models the platform itself.

\subsection{Step 3: Invariance and DIF with platform as the group}
\begin{figure}[t]
\centering
\includegraphics[width=0.8\linewidth]{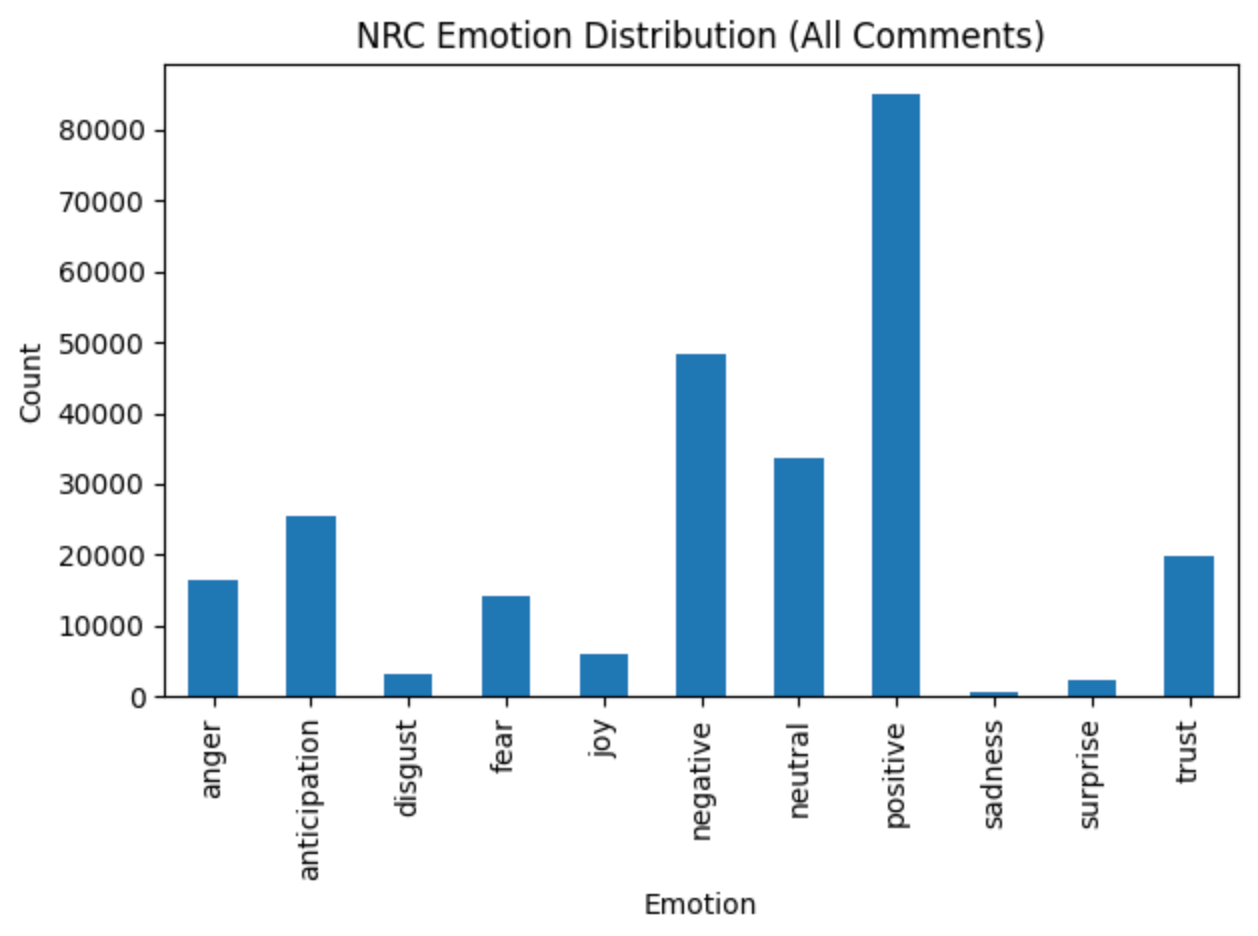}
\caption{Dissertation-stage NRC distribution over all comments (dissertation Figure~7.8, p.~40): eight emotion categories plus positive, negative, and neutral tallies. The bars sum to the classified corpus, so each comment carried one mutually exclusive label at dissertation stage; the application replaces this single-label assignment with binary presence codes per emotion, which become the Step~3 indicators.}
\label{fig:nrc}
\end{figure}

We fit multiple-group measurement models grouped by platform. The indicators are binary presence codes for each of the eight emotions of the NRC Emotion Lexicon \citep{mohammad2013} (Figure~\ref{fig:nrc}), derived from the raw lexicon match counts rather than a single dominant label (the executed notebook demonstrates this computation per comment; Figure~\ref{fig:nrcexample}), together with the three-category sentiment class treated as ordinal through a latent-response threshold model. Dimensionality is tested per platform before any invariance constraint is imposed, and the invariance battery is conditional on an acceptable configural model. Because the NRC indicators derive from a lexicon, they are excluded from the non-lexicon replication arm, which is restricted to the sentiment indicator set.

Four safeguards apply at this step. Categorical indicators enter through a latent-response threshold model, identified at each invariance level following \citet{wu2016} so that the tests do not depend on the identification constraints chosen for the baseline model; we do not treat categorical labels as linear indicators. Invariance decisions rest on effect-size criteria, changes in CFI and RMSEA \citep{chen2007, putnick2016}; chi-square values are reported but decisions do not rest on them, since at this sample size the test rejects trivial differences. Because these cutoffs were calibrated for continuous indicators under maximum likelihood, we treat them as heuristics subject to sensitivity analysis under the categorical estimator, and per-event DIF results aggregate into a platform-level claim by a rule fixed before estimation: an indicator-platform pair is flagged only when non-invariance appears in at least two events that pass the density audit, in both classifier arms; for the NRC emotion indicators, which exist only in the lexicon arm, the two-event rule applies within that arm and any flag is reported as lexicon-conditional. DIF is tested separately within each anchor event, so an event-specific artefact cannot be mistaken for a global platform effect. And because the choice of anchor items (the indicators held invariant to identify the model, as distinct from the anchor events) can itself build in the conclusion, we re-estimate DIF without pre-chosen anchor items \citep{chen2023} and run alignment optimisation \citep{muthen2018} alongside the constrained models. These are three identification strategies for the same model rather than independent tests, so agreement rules out artefacts of anchor-item choice and nothing more; a claim of non-invariance must survive all three, a claim of invariance must do the same, and disagreement is reported as inconclusive.

\subsection{Step 4: Multilevel adjustment}
Steps 2 and 3 could still confound measurement with composition. We therefore fit a multilevel model with posts nested in events and platform as a fixed effect. Time since event enters the model explicitly: platforms differ systematically in when their users respond, and this timing component may reflect real opinion dynamics as well as affordance. We therefore estimate platform contrasts with and without the latency term and report the pair, as total and direct platform differences, alongside rather than inside the measurement comparison. Where the multilevel structure permits, we also apply the cluster-bias test of \citet{jak2013}, which examines invariance violations across clusters directly.

\subsection{Classifier-artefact control and interpretive stance}
\begin{figure}[t]
\centering
\includegraphics[width=0.62\linewidth]{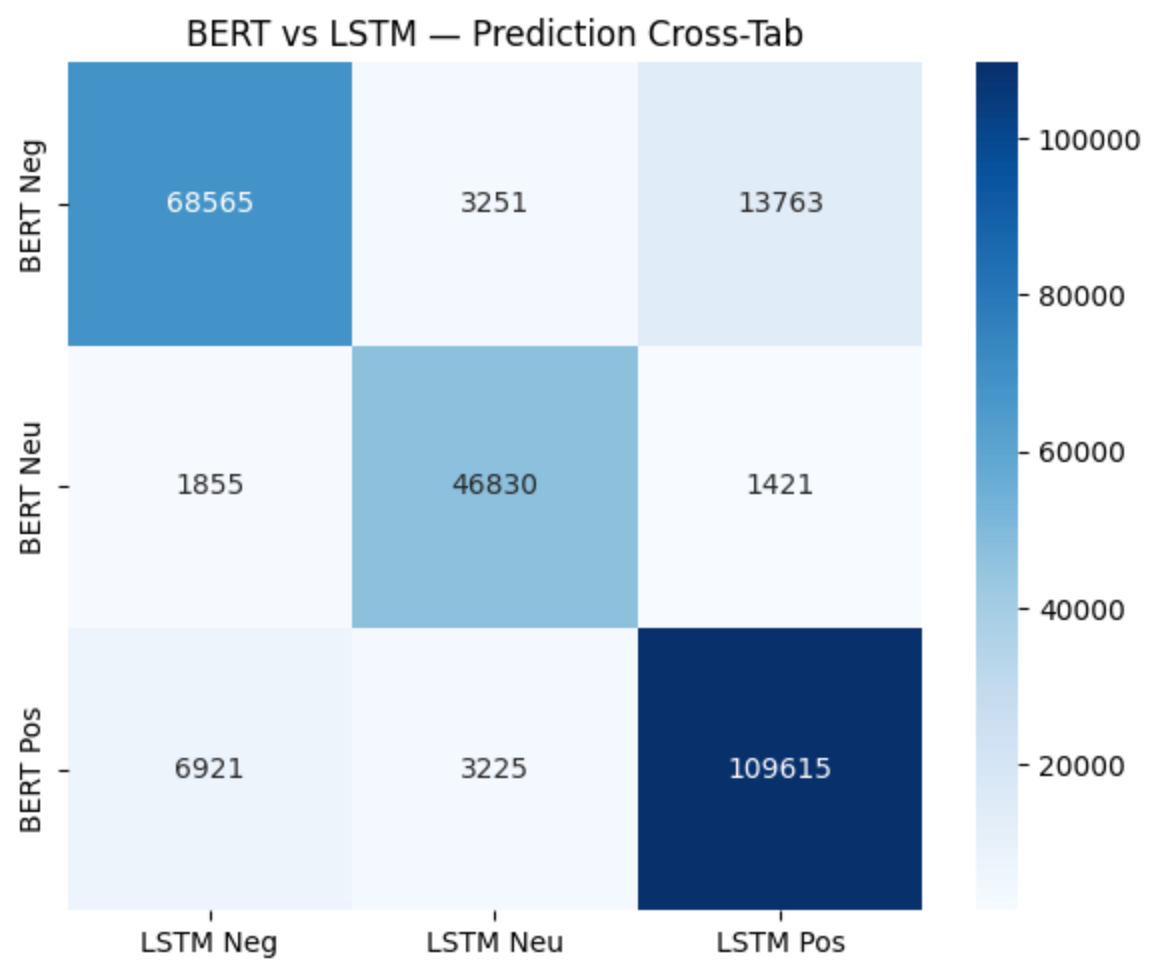}
\caption{Dissertation-stage cross-tabulation of BERT and BiLSTM predictions over the classified corpus (dissertation Figure~7.15, p.~43). Both families were trained on the same VADER-seeded labels, so their disagreement illustrates instrument-dependence within a shared label heritage; the BiLSTM is not the non-lexicon control, which is specified in the text.}
\label{fig:crosstab}
\end{figure}

\begin{figure}[t]
\centering
\includegraphics[width=0.75\linewidth]{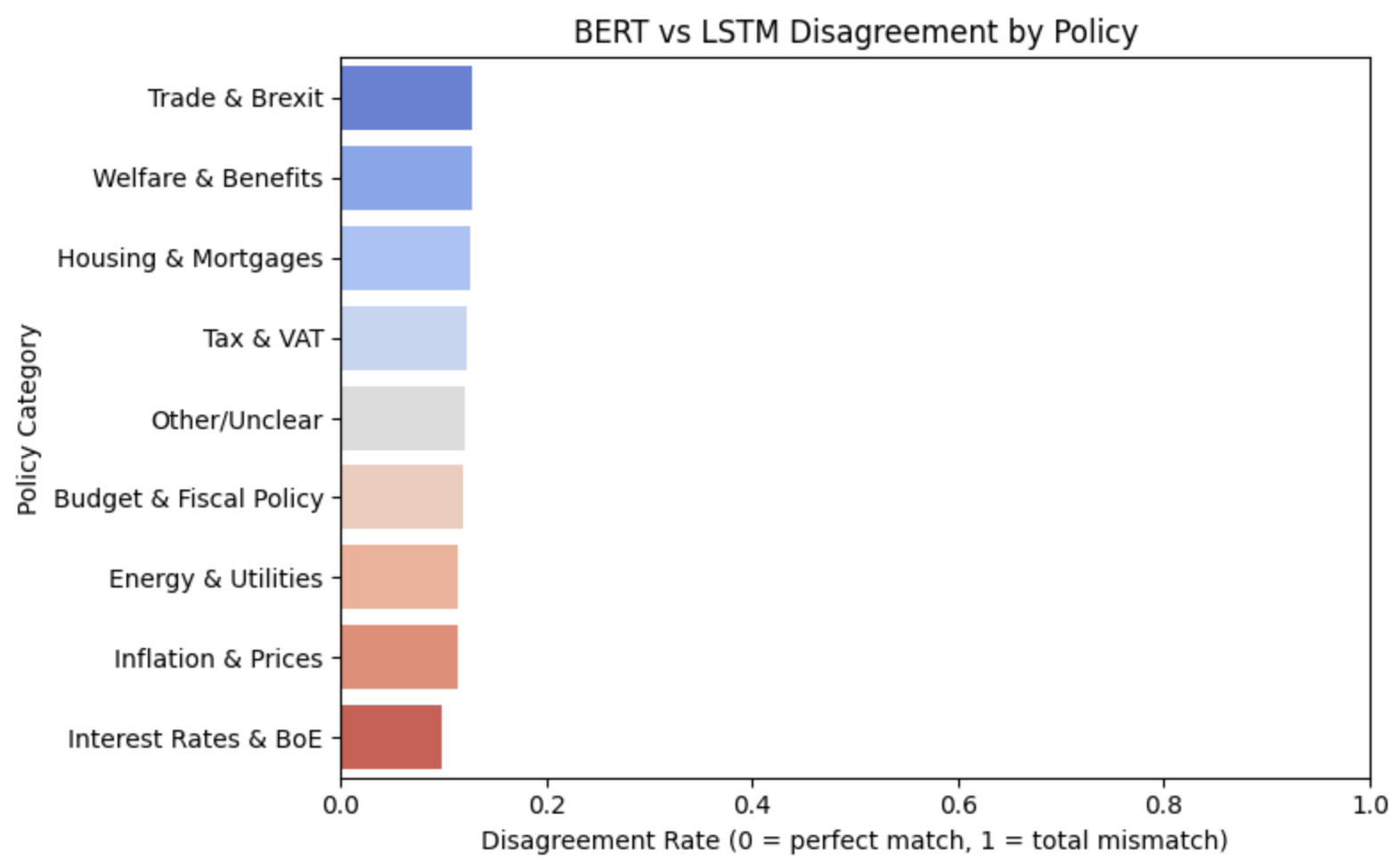}
\caption{BERT-versus-BiLSTM disagreement rate by policy category at dissertation stage (Figure~7.14, p.~43). Disagreement is highest for Trade and Brexit and Welfare and Benefits, though the spread across categories is modest, roughly 10 to 13 per cent.}
\label{fig:disagreement}
\end{figure}

\begin{figure}[t]
\centering
\includegraphics[width=\linewidth]{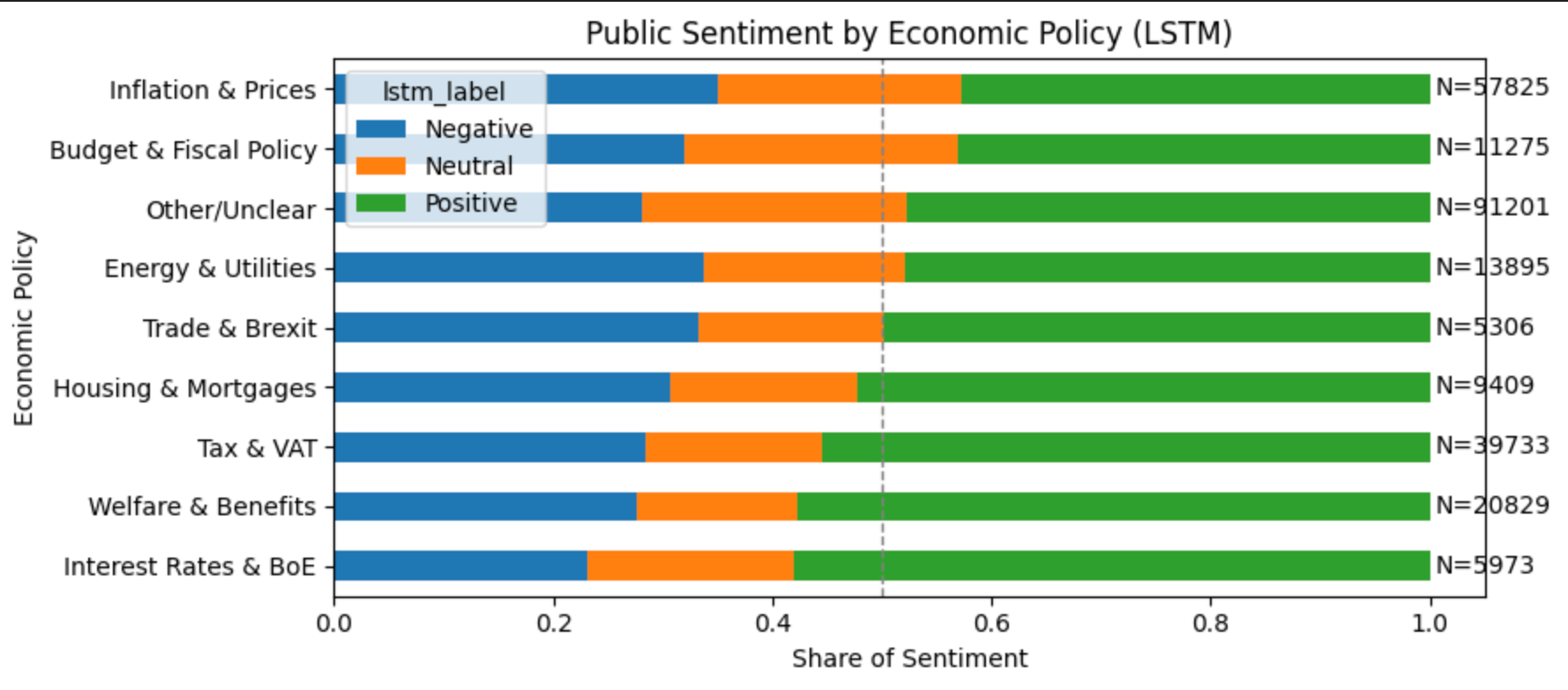}
\caption{The same corpus rescored with the BiLSTM (dissertation Figure~7.13, p.~42). The broad category ordering is nearly preserved, with adjacent categories swapping rank; measured shares are not preserved. Compare Figure~\ref{fig:policy}.}
\label{fig:policylstm}
\end{figure}

A platform effect estimated with one classifier could be an artefact of that classifier, and our training labels descend from a lexicon, as already disclosed. Every sentiment-indicator analysis is therefore replicated with a non-lexicon transformer, an off-the-shelf model fine-tuned on independently human-labelled sentiment data with no VADER ancestry (a Twitter RoBERTa of the kind released by CardiffNLP fits the requirement), and only convergent non-invariance, the same pattern in both classifier arms (the lexicon-seeded BERT and the non-lexicon transformer), is interpreted as platform-level non-equivalence rather than a lexicon artefact \citep{czarnek2022}. Convergence cannot rule out biases shared by both arms, a residual limitation stated in Section~\ref{sec:limitations}. Divergent patterns are reported as classifier artefacts, which also constrains the analyst degrees of freedom that make automated-annotation pipelines fragile \citep{baumann2025, atreja2025}. Classifiers disagree even when they share a label heritage: at dissertation stage BERT and the BiLSTM, both trained on the same VADER-seeded labels, agreed on 88.1 per cent of comments (225,010 of 255,446; executed notebook output, and Figure~\ref{fig:agreement}), which the dissertation text gives as 85 per cent, with disagreement varying modestly by topic, from roughly 10 to 13 per cent and highest for Trade and Brexit and for Welfare and Benefits (Figures~\ref{fig:crosstab} and~\ref{fig:disagreement}). This is why the protocol tests for convergence between arms rather than assuming it. The consequence for measurement is visible in Figure~\ref{fig:policylstm}: rescoring the same corpus with the BiLSTM nearly preserves the ranking of policy categories (only Energy and Utilities and Other/Unclear swap adjacent places) but changes the measured sentiment shares: with the corpus and its platform mix held fixed, the classifier alone moves the measurement, the same instrument-dependence this paper tests between platforms.

Finally, invariance evidence is diagnostic, not a licence \citep{robitzsch2023}. Passing a threshold does not certify pooled scores as interchangeable; it localises where comparability holds and where it fails, and every linking claim in Section~\ref{sec:linking} is conditioned on that localisation. The framework thereby formalises the platform-affordance error that \citet{sen2021} describe but do not test, and reaches a failure mode that label-error corrections do not touch, because those correct classifier error against a human criterion, not non-equivalence that the criterion itself shares across platforms \citep{egami2023}.

\section{Step 5: Linking}
\label{sec:linking}

If partial invariance holds, we link the platform scales. What we attempt is linking rather than equating: equating requires that transformed scores be interchangeable, a standard machine-labelled sentiment across platforms cannot meet, so we settle for the weaker relation \citep{kolen2014}. The linked object is a platform-level quantity, the platform-specific latent means and variances under the partial-invariance model, not any per-post score. The pre-identified fiscal events act as common stimuli rather than common items: different posts by different authors are scored on each platform, which in Kolen and Brennan's taxonomy is a moderation design, the weakest linking category, and we label it as such. The link is identified only under the assumption that latent sentiment towards each anchor event is equal in expectation across platform populations within the anchor window; where that assumption is doubted, the crosswalk removes population differences along with measurement differences, and we report both the naive pooled index and the adjusted index (Section~\ref{sec:cost}). The link also assumes that each platform's measurement function is stable across the anchor period, 2020 to 2025, during which platform affordances changed (Twitter's change of ownership in October 2022 among them); the per-event DIF estimates of Step 3 and the leave-one-event-out links are therefore reported in event order, so that a drift in the link, whether in measurement or in platform population, appears as a trend rather than being averaged into a single constant. The September 2022 mini-budget is the principal anchor, with the 48-hour and two-week windows defining the linking samples. Purification is ex ante: the rule for screening anchors for differential functioning is fixed before any linking function is estimated: the link is estimated, the anchor event with the largest standardised DIF statistic is removed, and the link is re-estimated until no statistic exceeds its threshold, with every removed event reported. Anchors therefore cannot be chosen in a way that favours a particular result. For sensitivity we re-estimate the linking function dropping each anchor event in turn and report the spread; a link that depends on a single event is reported as such; we do not smooth over it.

The protocol does not presuppose that linking will succeed. If Step 3 shows non-invariance too severe to support even partial-invariance linking, we do not link. The deliverable then becomes platform-specific sentiment series plus the demonstration that a pooled cross-platform index is unsupported for this construct. We would report either outcome: a crosswalk with quantified uncertainty, or evidence that none exists for this construct.

Linking functions, purification outcomes, and the leave-one-event-out sensitivity are outputs of the pre-specified application, conditional on its Steps 1 to 4.

\section{Step 6: The Cost of Assuming Equivalence in Simulation}
\label{sec:cost}

\begin{figure}[t]
\centering
\includegraphics[width=0.85\linewidth]{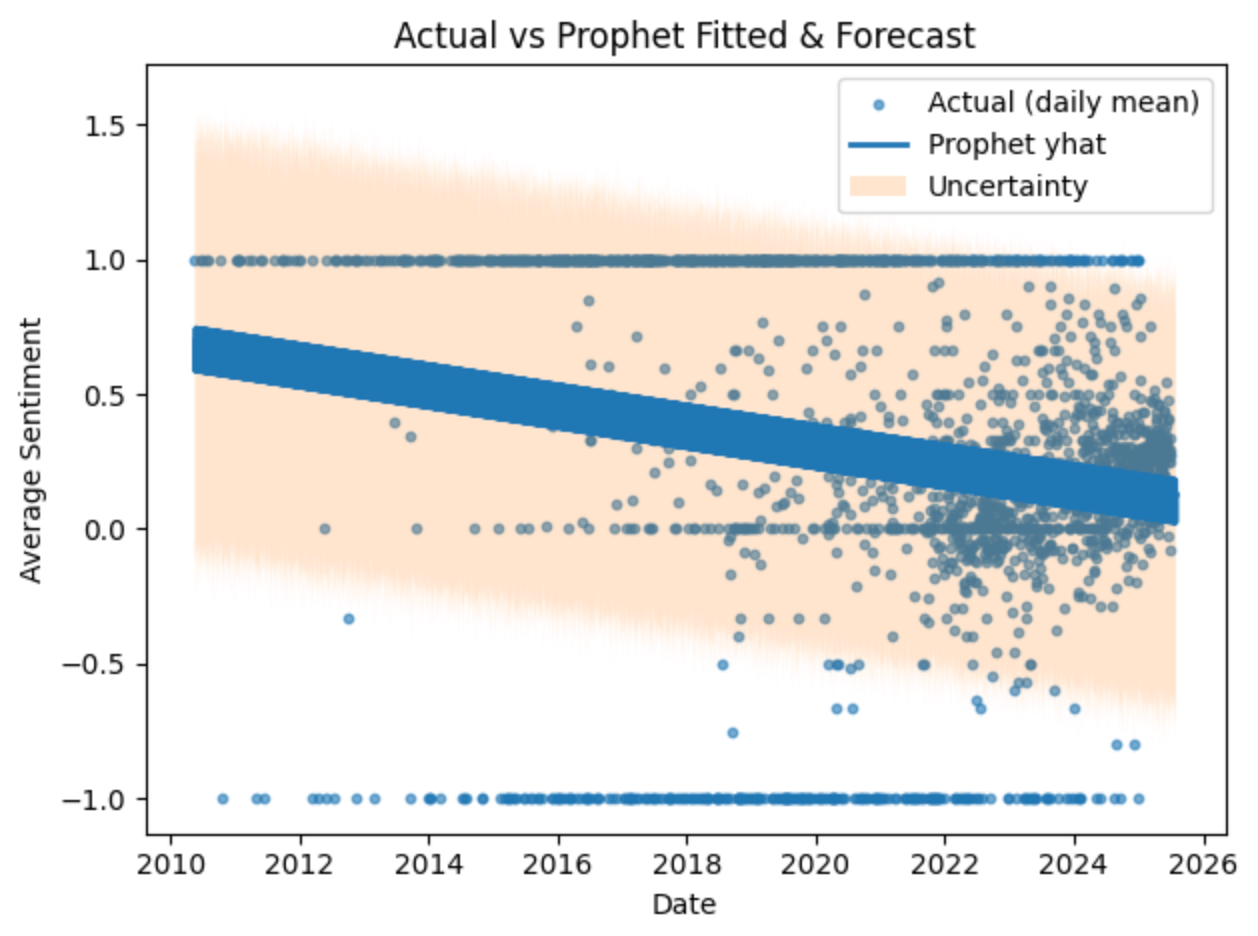}
\caption{The naive pooled index in practice: dissertation-stage daily mean sentiment pooled across platforms, with the Prophet fit and forecast band (dissertation Figure~7.7, p.~39). Several problems with the construction are visible here: days with a single post pin the daily mean at plus or minus one, no support threshold or volume weighting is applied, the uncertainty band ignores the bounded scale, and Twitter/X enters the pool only from January 2018 (dissertation p.~21), so the platform mix behind the daily mean shifts within the series. The simulation isolates the measurement component of exactly this construction.}
\label{fig:naive}
\end{figure}

\begin{figure}[t]
\centering
\includegraphics[width=0.8\linewidth]{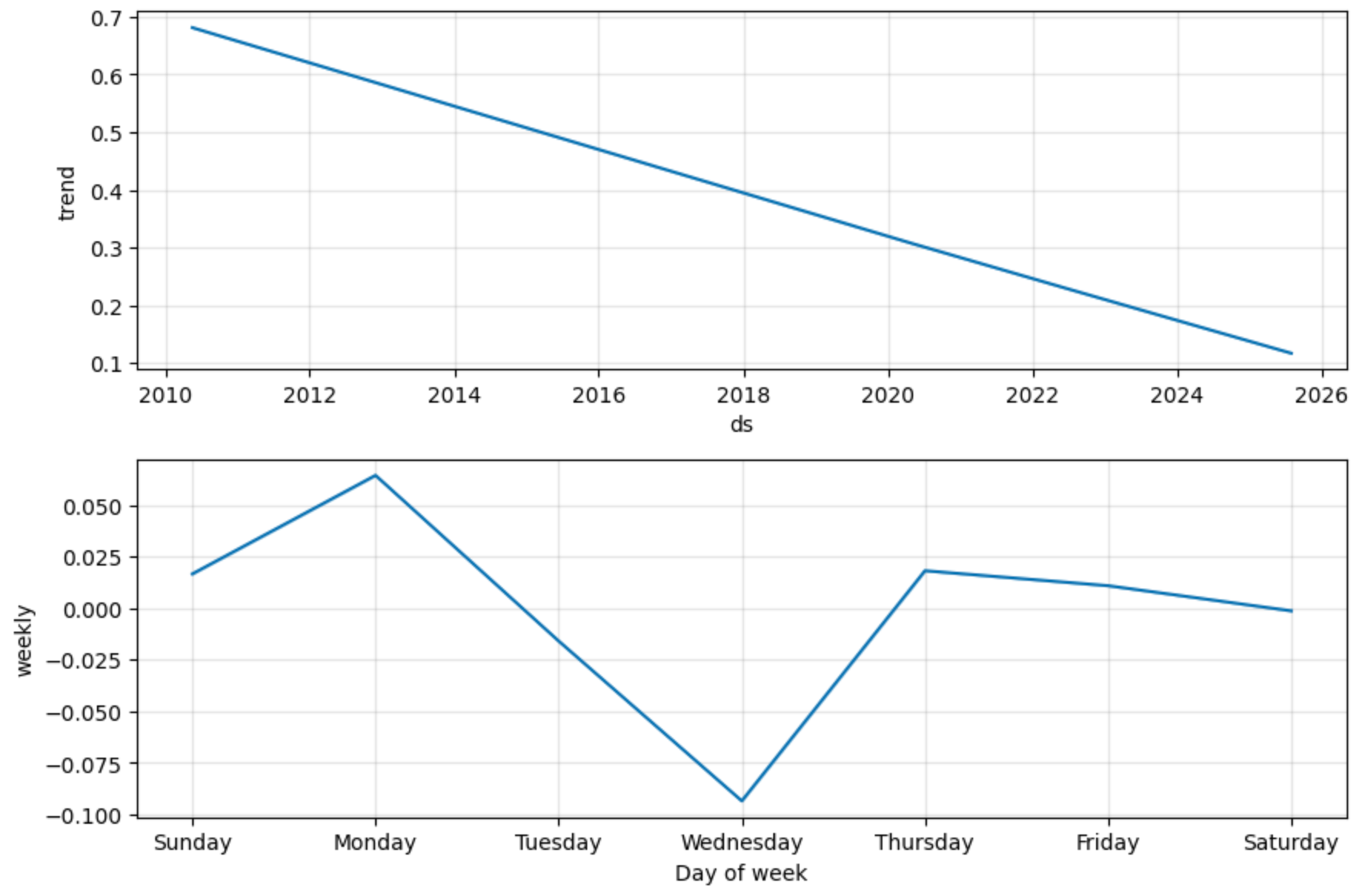}
\caption{Dissertation-stage Prophet decomposition of the pooled index (dissertation Figure~7.6, p.~39): a declining long-run trend and a midweek dip that we read as a candidate composition artefact, equally consistent with weekday differences in platform mix and with mood itself.}
\label{fig:seasonal}
\end{figure}

The final step quantifies the cost of assuming equivalence, an assumption the field usually makes without testing it. Unlike Steps 1 to 5, it does not require the corpus at all, because the ground truth is constructed: we simulate platforms whose measurement properties are known exactly, and ask how badly the standard index construction misreads them. The naive index pools posts across platforms, the construction canonical since \citet{oconnor2010}; the dissertation itself built such an index (Figure~\ref{fig:naive}), and its decomposition (Figure~\ref{fig:seasonal}) shows the trend and seasonal components an analyst would report from it.

\paragraph{Design.}
A latent daily sentiment series carries a known negative shift at an anchor event and recovers gradually. Each platform observes the same series through its own measurement function: an affordance bias shifts every expression, expression noise is added, and platform-specific thresholds discretise the result into negative, neutral, and positive; platform volume shares also drift modestly over time, a mild version of the composition change seen in the post-API record \citep{davidson2023, goanta2026}. Two indices are then computed daily. The naive index is the pooled mean of coded values across platforms. The adjusted index estimates each platform's latent value by inverting that platform's measurement function, then pools the estimates weighting each platform by its daily post volume. In the application the measurement parameters must be estimated, and a shift common to every expression on a platform is identified only relative to the other platforms, through the anchor assumption of Step 5 (Section~\ref{sec:linking}). In the simulation they are known, so the exercise isolates what adjustment can recover in principle and what pooling destroys.

\paragraph{Quantities reported.}
We compare the two indices on what an analyst would actually report: the estimated sentiment shift at the event and the day-by-day sign of aggregate sentiment; the recovery trajectory is shown for a single run in Figure~\ref{fig:simdemo}. The generating parameters were chosen to illustrate the mechanism and are not calibrated to any platform or corpus; they were fixed before the replications were run and are released with the code (Appendix~\ref{app:availability}). The event shift is estimated, for the latent series and for each index alike, as the mean over the ten days after the event minus the mean over the ten days before it; under the generating parameters, a step of $-0.5$ decaying with a ten day time constant, this contrast is $-0.332$. Across 500 replications (seeds 20261201 to 20261700), the naive index overstates it by $0.042$ on average (2.5th to 97.5th percentiles across replications, $0.032$ to $0.051$ in absolute size), a difference of roughly an eighth of the true shift that is a scale effect rather than an estimation error (see below), while the adjusted index recovers it with negligible bias (mean error $0.001$, under half of one per cent of the true shift).

The naive index also reports the wrong sign of aggregate sentiment, positive where the latent truth is negative or the reverse, on 13.8 of 60 simulated days on average (percentiles 11 to 17): for nearly a quarter of days, an analyst reading the pooled index would tell a policymaker that net sentiment points the wrong way. The adjusted index averages under one wrong sign day per run (mean 0.8, percentiles 0 to 2), so the naive count does not arise merely because the latent series sits near zero, although days near zero are where a biased index most easily changes sign (see below).

These figures are properties of the simulated configuration rather than bounds, and the two errors have different sources. The shift misstatement arises mainly from taking the discretised three point scale at face value: it persists, and is no smaller (a mean overstatement of $0.050$, against $0.042$ in the published configuration), when all platforms share one unbiased symmetric measurement function, so it is not a cost of non-equivalence. The wrong sign days all but vanish in that case and are specifically the cost of non-equivalence; their number grows with the spread between the platforms' measurement functions (a mean of 0.8, 1.8, 4.5, 13.8 and 42.9 wrong sign days at zero, a quarter, half, one and one and a half times the simulated spread) and depends on where latent sentiment sits relative to the platforms' net bias (with the pre-event baseline at $-0.1$ rather than $0.1$ the naive index makes no wrong sign days, and at $0.3$ about three), so the figure of nearly a quarter of days belongs to this configuration. Two further limits apply. Latent sentiment is identical on every platform by construction, so the simulation isolates the measurement component and is silent on the selection confound discussed in Section~\ref{sec:limitations}. And the adjusted index uses the true measurement functions; in the application those functions are estimated in Steps 3 and 5, and estimation error in them will cost the adjusted index some of its advantage, by an amount a later version of this paper will quantify by simulation and the follow-up study will report for the application.

\paragraph{What is deferred to the application.}
Corpus-based variants of this exercise, reweighting platform shares to mimic a platform losing API access, and the forecast-error comparison that feeds both indices to a Prophet model \citep{taylor2018} as the dissertation's forecasting stage did, require the application corpus and are pre-specified for the follow-up study. Corrections for label error \citep{egami2023} would not remove the simulated affordance bias: they align the classifier with its criterion, and that bias acts on the expressed text, which a human criterion reads too.

\section{Limitations}
\label{sec:limitations}

Our claims are subject to five limitations. First, the selection confound. Users choose their platforms, and the same-event design controls topic and time, not who is posting. Any non-invariance we detect therefore bundles population composition with measurement mode; because the two can offset as well as compound, the bundle cannot be read as a bound on the pure mode effect, only as the platform-level difference in how the construct is measured \citep{sen2021}. The convergence criterion \citep{czarnek2022} removes artefacts specific to one classifier arm from this bundle, though not biases the two arms share; it cannot remove selection. Separating the two would require random assignment of users to platforms, within-user cross-posting data, or user-level covariates that predict sentiment but are unaffected by platform, the social-media counterparts of the routes survey methodology uses to separate mode selection from mode measurement; we have none of these.

Second, there is the question of platform scope. We study four platforms, and the anchor-density audit may narrow the analysis to the subset with sufficient same-event coverage. All findings are conditional on the retained platforms and events. Exclusion of a sparse platform is a data limitation, not evidence that it measures equivalently. Likewise, following \citet{robitzsch2023}, we read invariance results as diagnostic evidence about this corpus, not as a general warrant to pool.

Third, the construct itself. Our instrument measures expressed sentiment, not policy support. Negative sentiment in a thread about inflation need not be opposition to any policy \citep{oconnor2010}. Our equivalence results concern the sentiment measure itself; any mapping to attitudes would need separate validation.

Fourth, a single country and language. All data are UK-focused and English. Machine scoring is language-dependent \citep{nogara2025}, and platform affordance mixes differ across national ecologies, so we make no cross-country claim.

Finally, provenance. Several dissertation-stage statistics are internally inconsistent or are not reproduced by the execution records: the corpus total (approximately 279,000 in the text against 255,446 in the pipeline records), the per-platform counts (summing to roughly 395,000; pp.~21-25), the post-calibration expected calibration error (0.05 in the methodology, 0.03 in the results; pp.~30, 36), the manual labelling coverage (0.07\% in one chapter and approximately 0.5\% in another; pp.~25, 45), the agreement rate between the two classifier families (approximately 85 per cent in the text against 88.1 per cent in the executed notebook), and the train-test split (temporal in the reproducibility appendix, stratified random 90/10 with seed 42 under newer libraries in the executed code). The dissertation-stage figures reproduced in this paper are included as evidence of the instrument as it stood; the application will produce fresh counterparts of every figure from its own corpus, and the dissertation's reliability diagrams, whose calibration stage is not recorded, are deliberately not reproduced. Every dissertation-stage statistic reported in this paper is either established by the execution records or attributed by page to the dissertation text, and is labelled accordingly; quantities those records do not settle, the per-platform composition and the collection window among them, are not estimated here. The empirical application proceeds by fresh collection under the pre-specified design, scoped to the dense anchor windows the design requires, with provenance logged from the first request.

\section{Ethics Statement}
\label{sec:ethics}

This is an observational study of public posts on public policy topics. The dissertation-stage study completed the University of Portsmouth online ethics review as a low-risk project (School of Computing; Certificate of Ethics Review TETHIC-2025-111094, with supervisor sign-off). The study complies with UK GDPR: no private, deleted, or direct-message content was collected; released artefacts contain no personal identifiers; and quoted examples are paraphrased. The dissertation-stage criterion set comprised 200 posts balanced across platforms, all labelled by hand by the first author, who reviewed each post's VADER-assigned label; the dissertation's Cohen's kappa of approximately 0.78 (p.~28) measures agreement between those hand labels and VADER, not between two annotators. The follow-up application's collection and annotation will carry their own ethics cover. The main foreseeable misuse is reading our diagnostics as a ranking of platforms by accuracy; they test equivalence and do not score quality, and we caution against that reading.

\section{Conclusion}
\label{sec:conclusion}

Until shown otherwise, the platform should be treated as part of the measurement instrument. We formalise that claim and specify its test, adapting the psychometric equivalence toolkit to machine-labelled policy sentiment with the platforms as the groups. The simulation shows what is at stake: across 500 replications of a single configuration fixed in advance, a pooled index that takes platform scores at face value reports the wrong sign of aggregate sentiment on nearly a quarter of days because the platforms measure through different functions, an error that an index adjusted with the known measurement functions all but avoids. Whatever the application then finds, something useful follows: if invariance holds we will know where pooling can be defended, and if it fails we will have an estimate of where, and by how much, pooling misleads.

\section*{Competing Interests} The first author is a co-founder and the chief technology officer of ArtemisAI (Artemis AI Limited), a social media analytics company. The company did not fund, host, or supply data for this work.

\bibliographystyle{plainnat}
\bibliography{references}
\appendix

\section{Robustness Battery}
\label{app:robustness}

Every headline result of the pre-specified application is paired with the checks below; the simulation of Section~\ref{sec:cost} reports replication intervals under its fixed generating parameters; a systematic sensitivity analysis over those parameters, and an arm in which the measurement parameters are estimated rather than known, are deferred to a later version of this paper.

\paragraph{Invariance and DIF.} Categorical-indicator models with threshold constraints, judged on effect-size criteria rather than chi-square. Alongside the constrained models we run alignment \citep{muthen2018} and anchor-free DIF \citep{chen2023}, so no conclusion depends on a hand-picked set of anchor items.

\paragraph{Classifier-artefact control.} The sentiment-indicator test suite is replicated in full with a non-lexicon transformer; the NRC emotion indicators, which derive from a lexicon, are assessed in the lexicon arm alone and any finding on them is labelled lexicon-conditional. A result counts as platform-level non-equivalence only when both classifier arms show it \citep{czarnek2022}; a pattern present under one arm alone is recorded as a classifier artefact.

\paragraph{Linking sensitivity.} Leave-one-event-out re-estimation of linking constants over the pre-identified fiscal-event anchors, with the September 2022 mini-budget as the main anchor; both 48-hour and 2-week event windows; anchor purification rule fixed before any estimation and applied iteratively (Section~\ref{sec:linking}).

\paragraph{Uncertainty and calibration.} Bootstrap confidence intervals on all indices; McNemar tests for classifier comparisons; temperature-scaled calibration checked per platform. At dissertation stage the executed notebook records McNemar $b=22{,}429$, $c=5{,}680$, $p<0.001$, and a bootstrap macro-F1 difference of 0.067 (95 per cent interval 0.065 to 0.068) between the two classifier families, against the VADER-seeded reference labels.

\paragraph{Annotation reliability.} Expansion of the human criterion sample from the dissertation's 200 balanced posts (all labelled by hand by the first author; the dissertation's Cohen's kappa of approximately 0.78 is agreement with VADER, not between annotators; dissertation p.~28) to a 1,000 to 1,500 comment probability sample with a second-annotator subset; the agreement statistics are outputs of the application.

\section{Dissertation-Stage Diagnostic Outputs}
\label{app:diagnostics}

\begin{figure}[!htbp]
\centering
\includegraphics[width=0.6\linewidth]{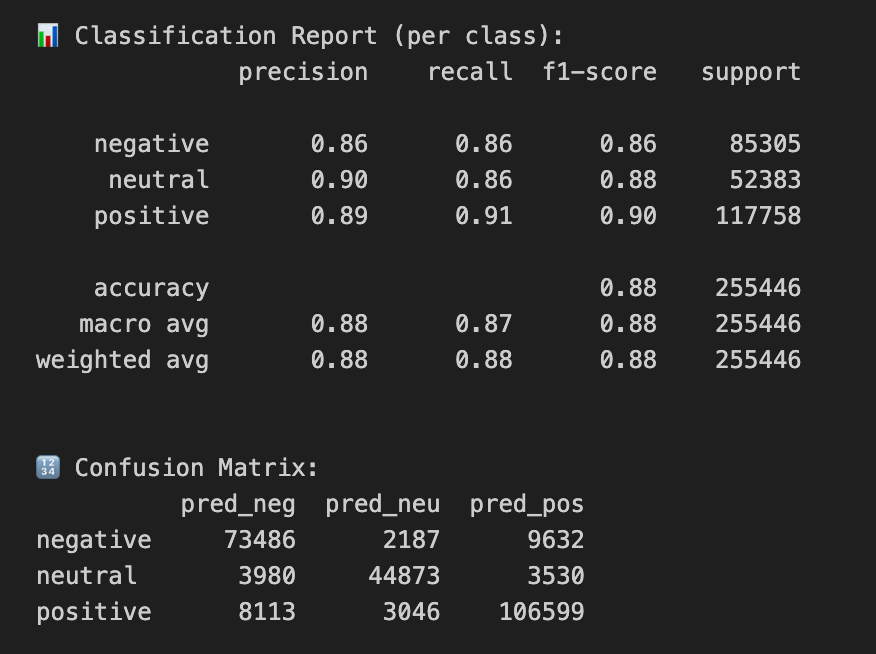}
\caption{Dissertation-stage per-class report and confusion matrix, fine-tuned BERT (dissertation Figure~7.1, p.~36). Class supports total 255,446; the reference labels are the VADER-seeded corpus labels, so these diagnostics measure agreement with the seed lexicon, not accuracy against the human criterion.}
\label{fig:confbert}
\end{figure}

\begin{figure}[!htbp]
\centering
\includegraphics[width=0.6\linewidth]{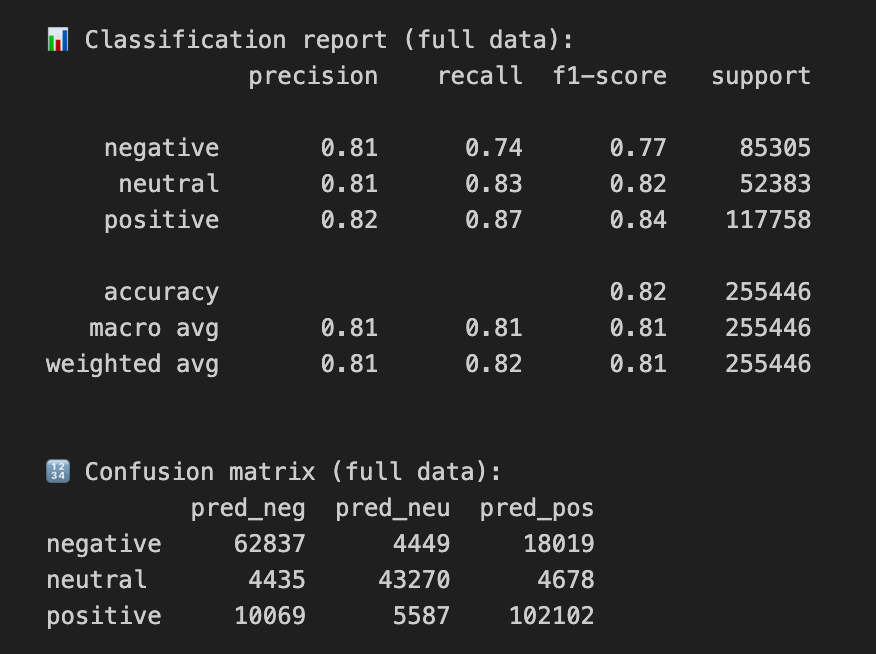}
\caption{Per-class report and confusion matrix for the BiLSTM at dissertation stage (Figure~7.2, p.~36). The reference labels are again the VADER-seeded corpus labels, not the human criterion.}
\label{fig:confbilstm}
\end{figure}

\begin{figure}[!htbp]
\centering
\includegraphics[width=0.55\linewidth]{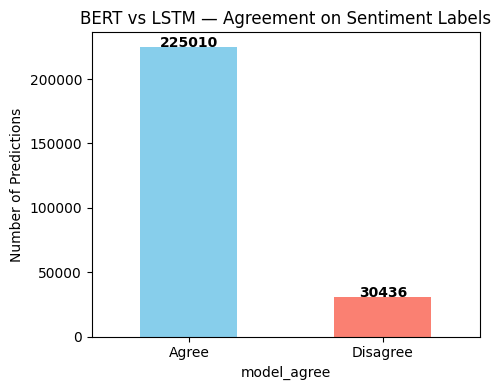}
\caption{Agreement between the two classifier families over the classified corpus, from the executed notebook: 225,010 of 255,446 comments (88.1 per cent). The dissertation text's figure of 85 per cent is superseded.}
\label{fig:agreement}
\end{figure}

\begin{figure}[!htbp]
\centering
\includegraphics[width=0.55\linewidth]{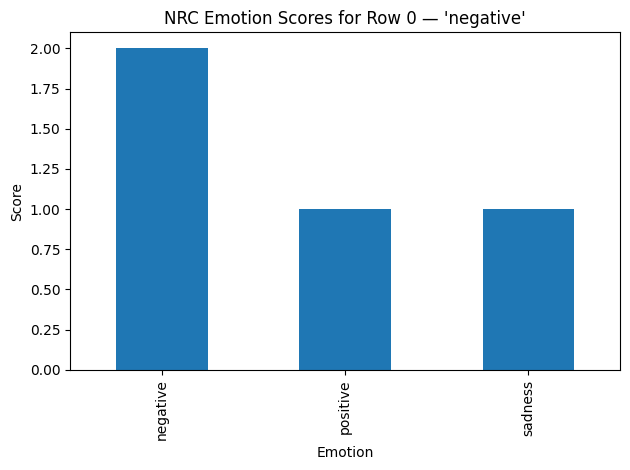}
\caption{Per-comment NRC lexicon match counts for one example (executed notebook). These raw counts, not the single dominant label, are the basis of the Step~3 binary emotion indicators.}
\label{fig:nrcexample}
\end{figure}

The dissertation-stage per-class diagnostics are reproduced here in their original form (Figures~\ref{fig:confbert} and~\ref{fig:confbilstm}), recording the state of the instrument at dissertation stage; they are console outputs rather than typeset figures; the application will produce typeset counterparts, with fresh reliability diagrams, from its own corpus. The application's figure standard is fixed now: console reports become typeset tables that name their reference labels; paired figures share category order; the BiLSTM is named consistently; legends sit outside the axes in plain English; unsigned rates use a sequential palette; axes are limited to the data range; interval levels are stated; output is vector. Against the VADER-seeded reference labels, both classifiers had their lowest F1 on the negative class (0.86 for BERT and 0.77 for the BiLSTM, against 0.88 and 0.82 on the neutral class), and their largest confusions were between negative and positive (Figures~\ref{fig:confbert} and~\ref{fig:confbilstm}); the dissertation text's statement that the neutral class was hardest (pp.~35-37) is not borne out by these outputs. The application's criterion sample will measure class-level performance against human labels. Figure~\ref{fig:agreement} shows the executed notebook's agreement tally between the two families, the source of the 88.1 per cent figure in Section~\ref{sec:framework}, and Figure~\ref{fig:nrcexample} an example of the per-comment NRC match counts from which the Step~3 binary indicators derive.

\clearpage
\section{Collection Keyword Scheme}
\label{app:keywords}

The dissertation's auditable seed rules (p.~18), reproduced here in typeset form; the pre-specified application fixes this scheme ex ante for its fresh collection. Six topics, each with a keyword list and, for the first two, regular-expression hints: taxation and fiscal policy (tax, VAT, income tax, NI, national insurance, HMRC, budget, chancellor, Finance Bill, fiscal, levy); inflation, cost of living and prices (inflation, CPI, prices, cost of living, food prices, fuel prices, energy bills, price cap, BoE, interest rate); welfare, benefits and Universal Credit (welfare, benefits, Universal Credit, UC, DWP, benefit cap, means-tested, child benefit); energy and utilities (energy price guarantee, Ofgem, electricity bill, gas bill, standing charge, fuel duty); housing and mortgages (mortgage, rent, landlord, housing benefit, stamp duty, help to buy); and the labour market (minimum wage, national living wage, pay rise, pay freeze, strike, unemployment). When more than one topic matched, priority ran taxation, then inflation, welfare, energy, housing, labour. Negation terms within three tokens of a polarity keyword flagged the post for review. A gazetteer of institutions (HM Treasury, Bank of England, DWP, Ofgem, OBR) and role mappings (Chancellor, Prime Minister, Shadow Chancellor) supplied entity hints.

\section{Data and Code Availability}
\label{app:availability}

The simulation code of Section~\ref{sec:cost}, with the generating parameters and seeds for every reported run and the script that draws Figure~\ref{fig:simdemo}, together with the anchor event list (Table~\ref{tab:anchors}) and the keyword scheme (Appendix~\ref{app:keywords}), is available at \url{https://github.com/asadfix/platform-measurement} and archived at \url{https://doi.org/10.5281/zenodo.23105129}. The application will add its annotation guidelines and aggregated per-event statistics. The dissertation-stage corpus and its labels are not redistributed: under the dissertation's ethics and data-management commitments no raw dataset was to be shared publicly and only anonymised, aggregated summaries were to be retained, and platform and licence terms precluded redistributing raw posts (dissertation pp.~15, 24, 25, 57). The dissertation stage ran on Google Colab (Python 3.11); the executed notebook's environment (torch 2.8, transformers 4.55, spacy 3.8, nltk 3.9, pandas 2.2) differs from the versions listed in the dissertation's reproducibility appendix (p.~19), and the application will pin and report its environment exactly. The executed split was stratified random, 90 per cent train and 10 per cent test (229,901 and 25,545 comments, seed 42), not the temporal split the reproducibility appendix describes; the application will use and report a temporal split alongside the random one. In the application, raw post text will not be redistributed, in line with platform terms, and post identifiers will be released only where terms permit rehydration. Released artefacts contain no personal data. Compute used: a single NVIDIA T4 GPU (16\,GB) on Google Colab Pro for the dissertation-stage models (dissertation p.~19); run times were not logged. The simulation runs in minutes on a laptop.

\end{document}